\documentclass[letterpaper, 10 pt, conference]{ieeeconf}  

\IEEEoverridecommandlockouts                              

\usepackage[utf8]{inputenc}
\usepackage{cite}
\usepackage{graphicx} 
\usepackage{balance}
\usepackage{booktabs}
\usepackage{multirow}
\usepackage{makecell}
\usepackage{url}
\usepackage{siunitx}
\title{\LARGE \bf
Stereo Multistage Spatial Attention for Real-Time Mobile Manipulation Under Visual Scale Variation and Disturbances
}

\author{Xianbo Cai$^{1}$,  Hideyuki Ichiwara$^{1,2}$, Hyogo Hiruma$^{1,3}$, Masaki Yoshikawa$^{1}$, Hiroshi Ito$^{1,3}$and Tetsuya Ogata$^{1,4}$%
\thanks{$^{1}$Xianbo Cai, Hideyuki Ichiwara, Hyogo Hiruma, Masaki Yoshikawa, Hiroshi Ito, and Tetsuya Ogata are with the Department of Intermedia Art and Science, Waseda University, Tokyo, Japan {\tt\footnotesize hourensou369@fuji.waseda.jp}}%
\thanks{$^{2}$Hideyuki Ichiwara is with SB Intuitions Corp., Tokyo, Japan. }%
\thanks{$^{3}$Hiroshi Ito and Hyogo Hiruma,is with Research and Development Group, Hitachi, Ltd., Ibaraki, Japan}%
\thanks{$^{4}$Tetsuya Ogata is with the National Institute of Advanced Industrial Science and Technology (AIST), Tokyo, Japan {\tt\footnotesize ogata.waseda.jp}}}

\begin{document}
\maketitle
\thispagestyle{empty}
\pagestyle{empty}

\begin{abstract}
Robots operating in open, unstructured real-world environments must rely on onboard visual perception while autonomously moving across different locations. Continuous changes in onboard camera viewpoints cause significant visual scale variations in target objects, affecting vision-based motion generation. In this work, we present a stereo multistage spatial attention-based deep predictive learning method for real-time mobile manipulation. The proposed methods extracts task-relevant spatial attention points from stereo images and integrates them with robot states through a hierarchical recurrent architecture for closed-loop action prediction. We evaluate the system on four real-world mobile manipulation tasks using a mobile manipulator, including rigid placement, articulated object manipulation, and deformable object interaction. Experiments under randomized initial positions and visual disturbance conditions demonstrate improved robustness and task success rates compared to representative imitation learning and vision-language-action baselines under identical control settings. The results indicate that structured stereo spatial attention combined with predictive temporal modeling provides an effective solution within the evaluated mobile manipulation scenarios.

\end{abstract}


\section{INTRODUCTION}
\begin{figure}[!t]
    \includegraphics[width=0.46\textwidth]{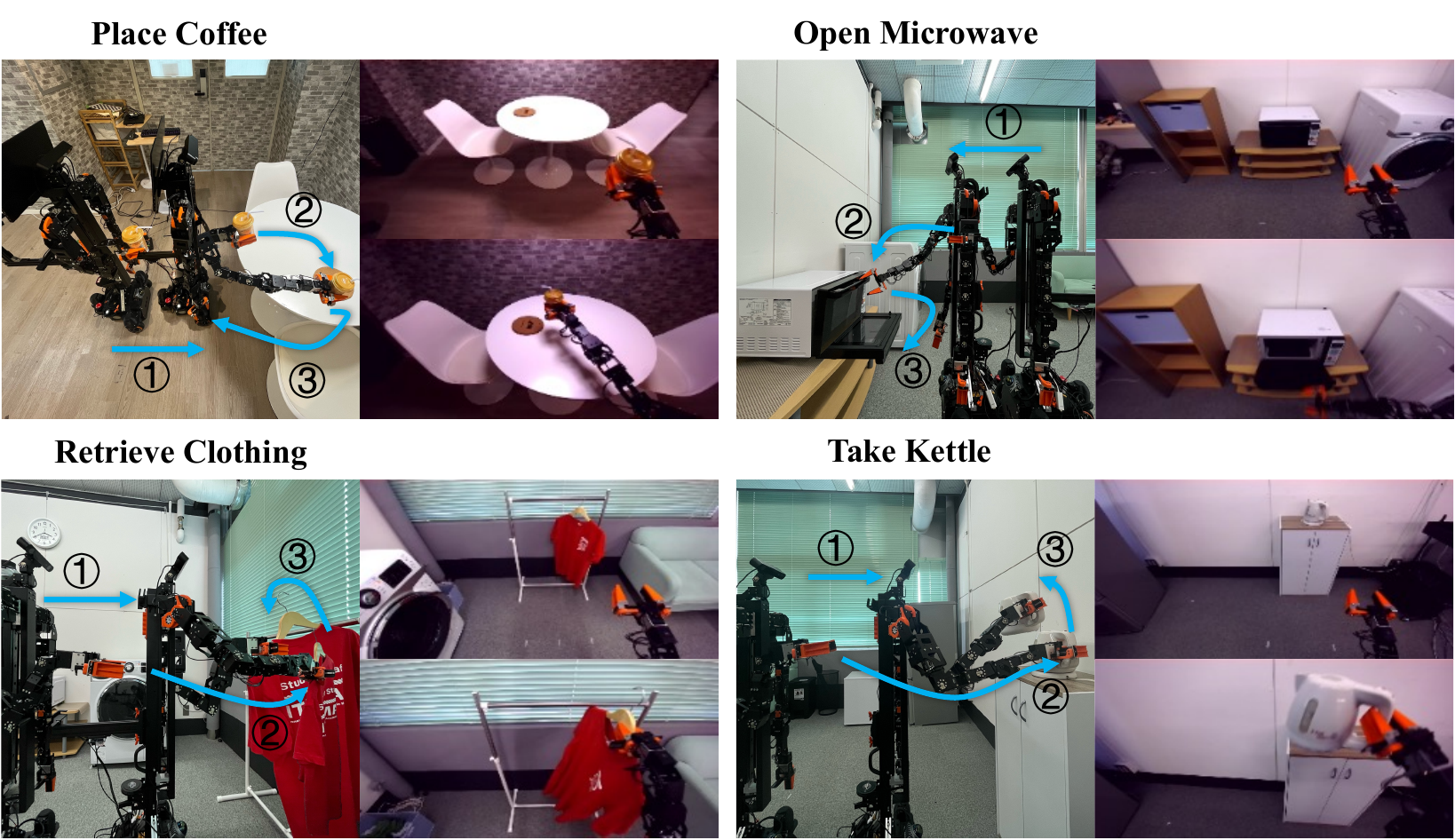}
    \caption{Examples of mobile manipulation tasks in this study. The blue lines indicate the robot’s movement, as well as the appearance of the target object in the onboard camera view.}
    \label{fig:Tasks}
\end{figure}

In recent years, there has been a growing demand for robots capable of operating autonomously in open, real-world environments such as homes, where they assist with daily tasks. In such unstructured settings, fixed external cameras are often unavailable, and robots must navigate autonomously across different locations to complete tasks. During this process, continuous changes in the onboard camera’s viewpoint cause significant variations in the apparent scale, position, and visibility of target objects. (Fig.~\ref{fig:Tasks}), which pose challenges to vision-based motion generation models by affecting their stability and performance \cite{tobin2017domain}\cite{zhu2023learning}.

Traditional methods based on explicit visual processing and geometric computation, such as feature point tracking \cite{lowe2004distinctive} and template matching \cite{fiala2005artag}\cite{wang2016apriltag}, can partially compensate for minor visual variations. However, they generally rely on accurate modeling of the environment or the target object. In contrast, deep learning-based methods have attracted increasing attention in recent years, particularly end-to-end imitation learning methods, which can automatically learn visual features and motion policies directly from data \cite{lenz2015deep}\cite{levine2018learning}. These methods have demonstrated impressive performance under controlled experimental conditions. However, they typically require large-scale training datasets, making data collection and training costly \cite{yu2020meta}. Also, the sim-to-real transfer problem remains a major challenge \cite{andrychowicz2020learning}, which may limit their applicability in real-time onboard control systems.

Deep predictive learning \cite{suzuki2023deep} has emerged as an imitation learning method with low training cost and strong robustness to environmental changes \cite{yang2016repeatable}\cite{ito2022efficient}\cite{ichiwara2022contact}. In particular, \cite{ichiwara2021spatial} introduced a spatial attention mechanism (SA) to enhance the extraction of object location features from images, significantly improving the model’s adaptability to state changes. However, most methods have been validated only in static environments with fixed robotic platforms, lacking systematic analysis of attention point stability and error accumulation in real-world mobile robot tasks.

In this work, we aim to build a practical mobile manipulation method that is robust to viewpoint-induced scale variations and visual disturbances while maintaining real-time closed-loop control. We propose a stereo multistage spatial attention-based deep predictive learning framework for end-to-end motion generation. The key idea is to extract task-relevant spatial attention points from stereo observations and integrate them into a hierarchical temporal prediction module for stable action generation. The proposed method incorporates: (1) A stereo multistage spatial attention (MSA) module that aggregates attention information across multiple feature stages to improve spatial consistency under viewpoint changes. (2) A hierarchical recurrent architecture that integrates stereo attention points and robot states for temporally coherent motion prediction. (3) A real-time closed-loop control implementation that runs efficiently on CPU hardware.

We conduct experiments across four real-world mobile manipulation tasks and compare the proposed method with representative methods, including ACT \cite{zhao2023learning}, Diffusion Policy \cite{chi2023diffusion}, as well as recent vision-language-action (VLA) models such as $\pi$0 \cite{black2024pi_0} and SmolVLA \cite{shukor2025smolvla}. Experimental results demonstrate that our method exhibits advantages in the following aspects: (1) Improve robustness to target size variations, (2) Tolerance to changes in initial distance, (3) Low inference latency, (4) Adaptability to visual disturbances. This work focuses on improving robustness within mobile manipulation scenarios subject to visual scale variation and moderate environmental perturbations. The results suggest that structured stereo spatial attention combined with predictive temporal modeling provides a practical solution for real-world mobile manipulation systems.

\section{RELATED WORK} 
\subsection{Traditional Vision-based and Rule-based Methods}
In mobile manipulation tasks, changes in camera viewpoints lead to visual scale variation of target objects, challenging vision-based motion generation methods. Traditional methods such as SIFT \cite{lowe2004distinctive} and ORB \cite{rublee2011orb} offer limited invariance to scale and rotation but often fail under large viewpoint shifts or poor texture. Template-based methods like ArTag \cite{fiala2005artag} and AprilTag2 \cite{wang2016apriltag} provide fast pose estimation but require extensive template sets and struggle with occlusion and lighting changes. In rule-base methods, Parosi et al. \cite{parosi2023kinematically} combined kinematically decoupled impedance control with image‑based visual servoing (IBVS). This method allows a quadruped mobile platform to maintain stable tracking under severe scale changes. However, it depends heavily on continuous visual feature tracking and degrades when features leave the camera view or become occluded. González‑Huarte et al. \cite{gonzalez2024visual} developed a position‑based visual servoing (PBVS) architecture with a finite‑state machine that continuously estimates the 3D pose of moving targets to achieve high‑precision manipulation. This method requires high‑quality depth sensing and accurate pose estimation, making it less robust in environments with occlusion. 

\subsection{Learning-based Methods}
In reinforcement learning (RL), Herzog et al. \cite{herzog2023deep} achieved high generalization and success rate on a mobile waste sorting task using multiple mobile robotic arms with up to 9527 hours of online RL training in a office building. However, like other reinforcement learning methods, they suffer from high training costs, dependence on the collection environment, and the generalization limited to specific task.

End-to-end deep learning methods offer more flexibility by automatically learning visual features and motion policy from expert data. For example, Dex-Net 2.0 \cite{mahler2017dex} uses a convolutional neural network to estimate grasp success probability and pose from a single depth image. Its effectiveness comes from training on millions of synthetic samples generated with randomized object shapes, poses, and camera views. As a result, the model learns visual patterns related to grasp success and achieves robustness under moderate viewpoint variations. However, Dex-Net 2.0 focuses on static single-view grasping, and its high training cost and sim-to-real gap limit application.

Imitation learning methods such as Action Chunking with Transformer (ACT) \cite{zhao2023learning} and Diffusion Policy (DP) \cite{chi2023diffusion} have shown promising results in motion generation tasks. ACT utilizes transformer architectures with action chunking to efficiently capture temporal dependencies and multimodal input relationships, showing high performance in specific manipulation tasks. DP leverages diffusion models to multimodal action distributions through iterative denoising processes. These approaches have shown strong performance in manipulation tasks, particularly under controlled experimental conditions. Recent vision-language-action (VLA) models for robotics, such as $\pi$0 \cite{black2024pi_0} and SmolVLA \cite{shukor2025smolvla}, are trained on large-scale, heterogeneous robotic data and demonstrate promising cross-task and cross-robot generalization. However, their performance on manipulation tasks still relies on task-specific fine-tuning, and their large model sizes typically lead to high inference costs.

Deep predictive learning based on the principle of free energy minimization \cite{friston2005theory}, is an end-to-end motion generation method that uses recurrent neural networks (RNNs) to learn the relationship between visual and motion data across time steps. During execution, the robot can predict future perceptual and motor data and adapt to unknown environments by minimizing the error between predicted and reality. This method avoids complex environment modeling and significantly reduces the costs. Several studies \cite{ito2022efficient}\cite{saito2021select}\cite{hiruma2022deep} have demonstrated its effectiveness and real-world motion generation adapted to changes. Notably, Ichiwara et al. \cite{ichiwara2021spatial} introduced a spatial attention mechanisms (SA) to enhance the extraction of target objects location in images, thereby improving robustness to changes. Most existing models are limited to static single-view setups.

\subsection{Stereo Vision and Disparity-Based Perception}
Stereo vision is increasingly used for manipulation tasks because of being able to provide disparity-based depth information. Liu et al. \cite{liu2020keypose} demonstrated the potential of stereo vision by applying weight-sharing CNNs to accurately estimate transparent object poses using large-scale manually annotated datasets. Cai et al. \cite{cai20243d} utilize stereo spatial attention with deep predictive learning, enabling the robot to infer 3D spatial positions of transparent objects and generate corresponding grasping motions. 

\begin{figure*}[!t]
    \centering
    \includegraphics[width=0.93\textwidth]{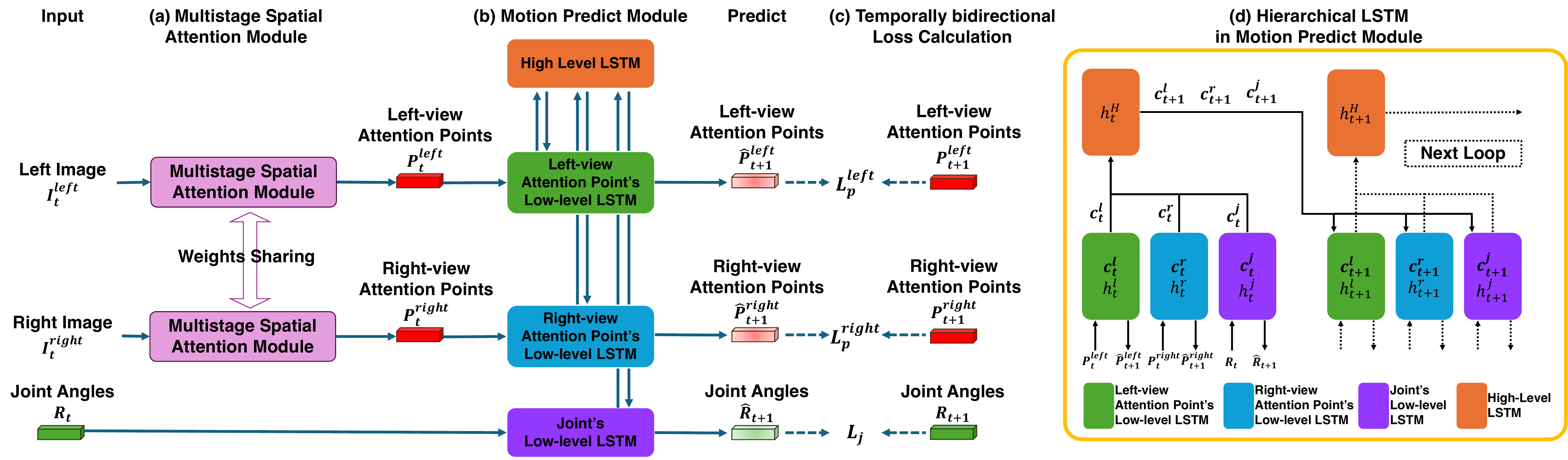}
    \caption{The overview of the proposed method: (a) multistage spatial attention module, (b) motion predict module, (c) temporally bidirectional loss, (d) hierarchical LSTM in motion predict module.}
    \label{fig:overview}
\end{figure*}

\section{PROPOSED METHOD}
The proposed method comprises a stereo multistage spatial attention module for task-relevant attention point extraction and a hierarchical LSTM for motion generation (Fig.~\ref{fig:overview}). The model's input includes the left $I^{left}_t$ and right $I^{right}_t$ camera images, and the robot motor data $R_t$ at time $t$. The outputs are robot motor data $\hat{R}_{t+1}$ and attention points $\hat{P}_{t+1}$ at the next time $t+1$. The model achieves real-time robot motion generation by repeatedly predicting the next robot motor data and applying them to the robot \cite{ito2022efficient}\cite{ichiwara2021spatial}\cite{saito2021select}\cite{ hiruma2022deep}. 

\begin{figure}
    \centering
    \includegraphics[width=0.45\textwidth]{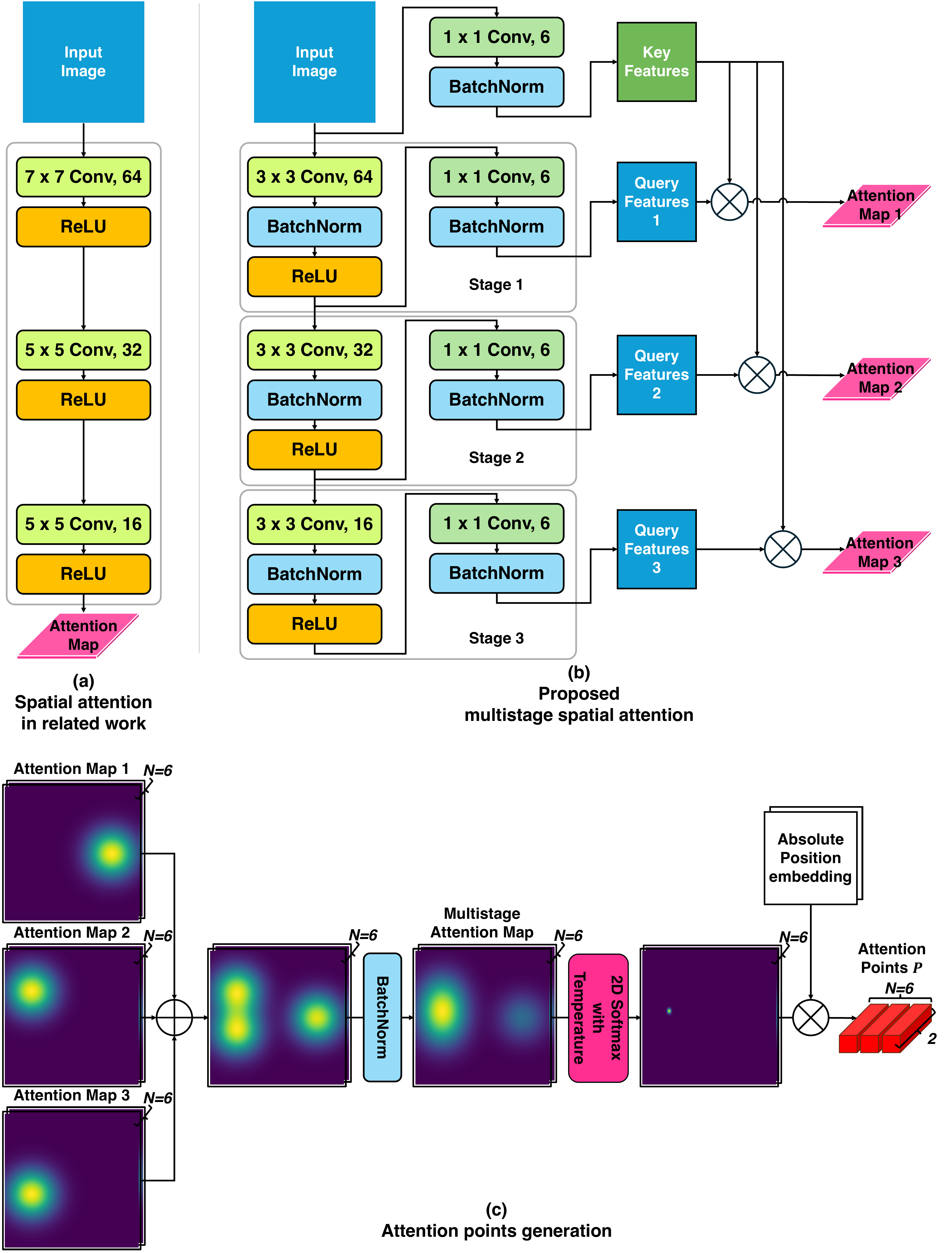}
    \caption{Detailed architecture of the spatial attention module: (a) spatial attention in related work, (b) proposed multistage spatial attention, (c) attention points generated from the attention maps produced by the proposed multistage spatial attention.}
    \label{fig:about_msa}
\end{figure}

\subsection{Stereo Spatial Attention Module}
The stereo multistage spatial attention (MSA) module (Fig.~\ref{fig:about_msa}) extracts attention points for motion generation. The stereo images are processed through weight-shared MSA modules, ensuring that the extracted attention points correspond to the same target objects. We set the MSA to extract $N=6$ attention points to ensure attention to task-relevant objects. The MSA extracts attention points $P_t^{left}$ and $P_t^{right}$ from the stereo images $I_t^{left}$ and $I_t^{right}$ at timestep $t$.

\subsubsection{Multistage feature extraction} 
Previous spatial attention (SA) \cite{lenz2015deep}\cite{ichiwara2022contact}\cite{ichiwara2021spatial} used three CNN layers to generate the attention maps solely from the final feature stage, forming a single-stage process (Fig.~\ref{fig:about_msa}(a)). When the input image passes through these CNN layers, task-relevant regions, such as the target object or the robot gripper, gradually become more salient, with their values correspondingly increasing.

In contrast, our method employs a three-stage CNN architecture that extracts image features at multiple scales (Fig.~\ref{fig:about_msa}(b)). The image features obtained from a linear transformation of the input image, which preserve the original spatial information, are defined as the $Key$ features, whereas the features produced by each CNN stage, which contain task-relevant representations, are defined as the $Query$ features. The outputs of all CNN stages are projected into $Query$ features using a learned linear channel projection (1×1 convolution) followed by BatchNorm2D, while the input RGB image is projected into a $Key$ feature using the same operations. Each stage adopts 3×3 convolutions, BatchNorm2D, and ReLU activations to achieve robust and efficient nonlinear feature extraction. All $Query$ and $Key$ features are projected to the same channel dimension ($=N$), ensuring consistent spatial resolution across stages. This multistage design enables the model to capture both fine-grained and coarse visual information relevant to the task.

\subsubsection{Cross-stage attention} For each stage, a Query feature is matched with the Key feature through a dot-product attention operation: 

{\scriptsize
\begin{equation}
    Attentionmap = \frac{Query\cdot Key}{\sqrt{H_{img}W_{img}}}
    \label{eq:attention map}
\end{equation}
}

where $H_{img}$ and $W_{img}$ denote the input image height and width. Like the scaling term stabilizes gradients in\cite{vaswani2017attention}, we apply spatial normalization to stabilize gradients. This produces three attention maps corresponding to a different visual scale.
  
\subsubsection{Multistage attention map fusion} The three attention maps are averaged and followed by BatchNorm2D to normalize response distributions. This normalizes the fused attention map and produces a stable attention representation (Fig.~\ref{fig:about_msa}(c)).
{\scriptsize
\begin{equation}
A_{fused}=BatchNorm2D\left( \frac{1}{3}\sum_{s=1}^{3}Attention~map_s\right)
\end{equation} 
}

\subsubsection{Attention point generation} To isolate the task-relevant locations, we apply a 2D softmax with temperature $T$, which sharpens the distribution and suppresses irrelevant peaks. The resulting map is then multiplied with absolute positional embeddings $PE_{h,w}$ to compute the 2D attention points $P$, which is a differentiable soft-argmax operation:

{\scriptsize
\begin{equation}
    PE_{h,w} = \left( \frac{h}{H_{\mathrm{img}}-1},\; \frac{w}{W_{\mathrm{img}}-1} \right),
    \label{eq:absolute_positional_embedding}
\end{equation}
}
\vspace{-6pt}
{\scriptsize
\noindent  where $0 \le h < H_{\mathrm{img}}$ and $0 \le w < W_{\mathrm{img}}$
}

{\scriptsize
\begin{equation}
    P = Softmax2D\left( \frac{A_{fused}}{T} \right) \odot PE_{h,w}, 
    \qquad T = 0.001
    \label{eq:attention_points}
\end{equation}
}

Repeating this process for all $N$ channels yields the stereo task-relevant attention points used by the motion predict module. The stereo configuration provides complementary viewpoint information that improves spatial consistency \cite{cai20243d}.

\subsection{Robot Motion Predict Module}
The Motion Prediction Module (Fig.~\ref{fig:overview}(b)) learns the temporal relationship between the robot’s motor data and attention points. To model these multimodal temporal dependencies, we employ a hierarchical LSTM architecture (Fig.~\ref{fig:overview} (d)). This structure allows the model to process different types of input signals at appropriate levels of abstraction, enabling feedback from the global context.

At each time step $t$, three low-level LSTMs independently process: the left-view attention points $P_{t}^{left}$, the right-view attention points $P_{t}^{right}$ and the robot’s motor data $R_{t}$. For each modality, a low-level LSTM updates its hidden state $h_{t}$ and cell state $c_{t}$: ($h_t^{l}$, $c_{t}^{l}$), ($h_t^{r}$, $c_{t}^{r}$) and ($h_t^{j}$, $c_{t}^{j}$), corresponding to left-view, right-view, and robot's motor data, respectively. Each low-level LSTM predicts the next-step values. To integrate information across modalities, the cell states from the three low-level LSTMs are concatenated and input into a high-level LSTM. The high-level LSTM updates its states:

{\scriptsize
\begin{equation}
    (h_{t+1}^{H}, [c_{t+1}^{l}, c_{t+1}^{r}, c_{t+1}^{j}]) = LSTM_{H}(h_{t}^{H}, [c_{t}^{l}, c_{t}^{r}, c_{t}^{j}])
    \label{eq:hlstm}
\end{equation}
}

This hierarchical interaction stabilizes long-horizon motion prediction and aligns the predicted attention trajectories with the predicted robot motion.

Eq.\ref{eq:total loss} defines the loss function $L$, combining the prediction mean squared error of robot joint state $L_j$ and attention point $L_p$. A smoothing loss is added to $L_j$ to enhance the motion continuity (Eq.\ref{eq:lj loss}). Without ground-truth labels for attention points, we use temporally bidirectional loss (Fig.~\ref{fig:overview} (c)) by comparing MSA-extracted attention points at $t+1$ with hierarchical LSTM predictions at $t+1$, yielding a loss (Eq.\ref{eq:lp loss}) to stabilize attention point learning. The weight coefficient $\alpha$ for $L_p$, starting at 0.0001 and increasing to 0.1 during training, balances exploration of task-relevant objects early on and refinement in later stages.

{\scriptsize
\begin{equation}
    L = L_j + \alpha Lp
    \label{eq:total loss}
\end{equation}
}

{\scriptsize
\begin{equation}
    L_j = \frac{1}{n} \sum_{i=1}^{n}\left\{ \left( \hat{R}_{t+1}^{i} - R_{t+1}^{i}\right)^2 + 0.1*\left(\hat{R}_{t+1}^{i} -\hat{R}_{t}^{i}\right)^2 \right\}
    \label{eq:lj loss}
\end{equation}
}

{\scriptsize
\begin{equation}
    L_p = \frac{1}{m} \sum_{k=1}^{m}\left\{ \left( \hat{P}_{t+1}^{left^{k}} - P_{t+1}^{left^{k}}\right)^2 +\left(\hat{P}_{t+1}^{right^{k}} - P_{t+1}^{right^{k}}\right)^2 \right\}
    \label{eq:lp loss}
\end{equation}
}


\begin{figure}[h!]
    \centering
    \includegraphics[width=0.48\textwidth]{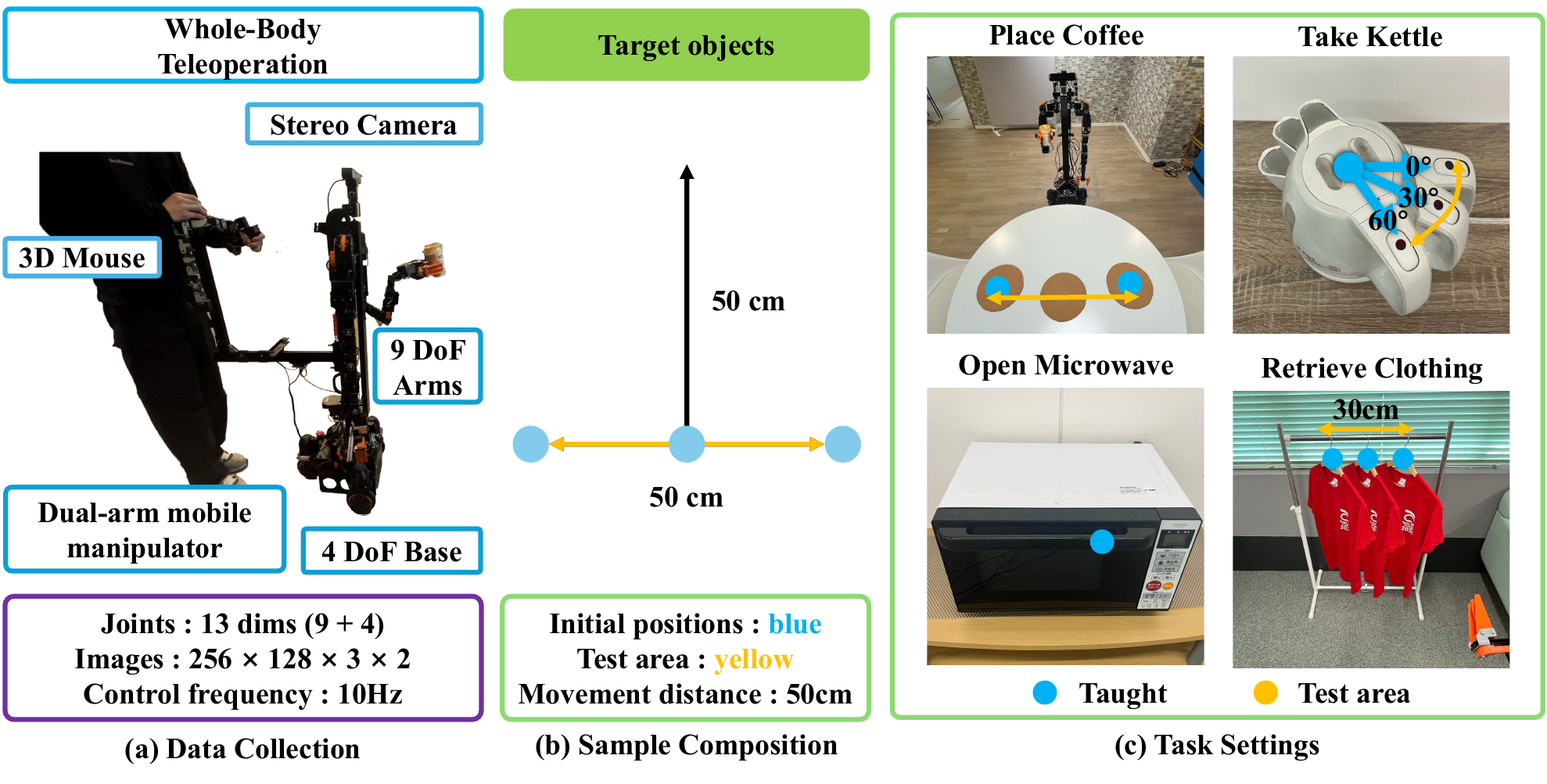}
    \caption{Experimental setup.}
    \label{fig:Setup}
\end{figure}

\begin{table}[h!]
\centering
\caption{Training Settings Across All Methods}
\label{tab:training_settings}
\resizebox{0.45\textwidth}{!}{
\renewcommand{\arraystretch}{1.0}
\setlength{\tabcolsep}{3pt}
\begin{tabular}{lccccc}
\toprule
\textbf{Model} &
\makecell{\textbf{Training}\\\textbf{steps}} &
\makecell{\textbf{Batch}\\\textbf{size}} &
\textbf{Optimizer} &
\makecell{\textbf{Learning}\\\textbf{rate}} &
\makecell{\textbf{Weight}\\\textbf{decay}} \\
\midrule

MSARNN (SA) &
50K &
4 &
Adam &
1e-4 &
1e-5 \\

\textbf{MSARNN (MSA)} &
50K &
4 &
Adam &
1e-4 &
1e-5 \\

ACT &
100K &
8 &
AdamW &
1e-5 &
1e-4 \\

Diffusion Policy &
200K &
32 &
Adam &
\phantom{0}1e-4$^{*}$ &
1e-6 \\

SmolVLA (0.45B) &
30K &
32 &
AdamW &
\phantom{0}\phantom{0}1e-4$^{**}$ &
\phantom{0}1e-10 \\

$\pi0$ (3.5B) &
30K &
32 &
AdamW &
\phantom{0}2.5e-5$^{**}$ &
1e-2 \\
\bottomrule
\end{tabular}
}
\vspace{1pt}
\begin{minipage}{0.45\textwidth}
\scriptsize
\centering
$^{*}$ With cosine decay (500 warmup).\\
$^{**}$ With cosine decay (1k warmup, 30k decay, final lr $2.5\times10^{-6}$).
\end{minipage}
\end{table}

\section{EXPERIMENTS}
To validate the effectiveness of proposed model, we use a dual-arm mobile manipulator for real-world experiments. This robot has two 9-DoF arms, a omni-directional mobile base with independent 3-wheel steering. A ZED stereo camera is mounted on the head and the camera angle remains fixed. The dataset is collected by a teleoperation system (Fig.~\ref{fig:Setup} (a)). Sequences are recorded at 10 Hz over 15 s, including the RGB stereo images (3×128×256×2), 9-DoF right arm, and 4-DoF mobile base motor data. The robot is taught to move 50 cm forward from one of three initial positions (Fig.~\ref{fig:Setup} (b)). For each task we collected 54 successful demonstrations. Multiple taught positions are set at the target object to increase the data variety. We evaluated the model in four mobile manipulation tasks (Fig.~\ref{fig:Setup} (c)). In each task, 50 tests were performed with randomized both the initial object and robot position.

\begin{figure*}[!t]
    \centering
    \includegraphics[width=0.95\textwidth]{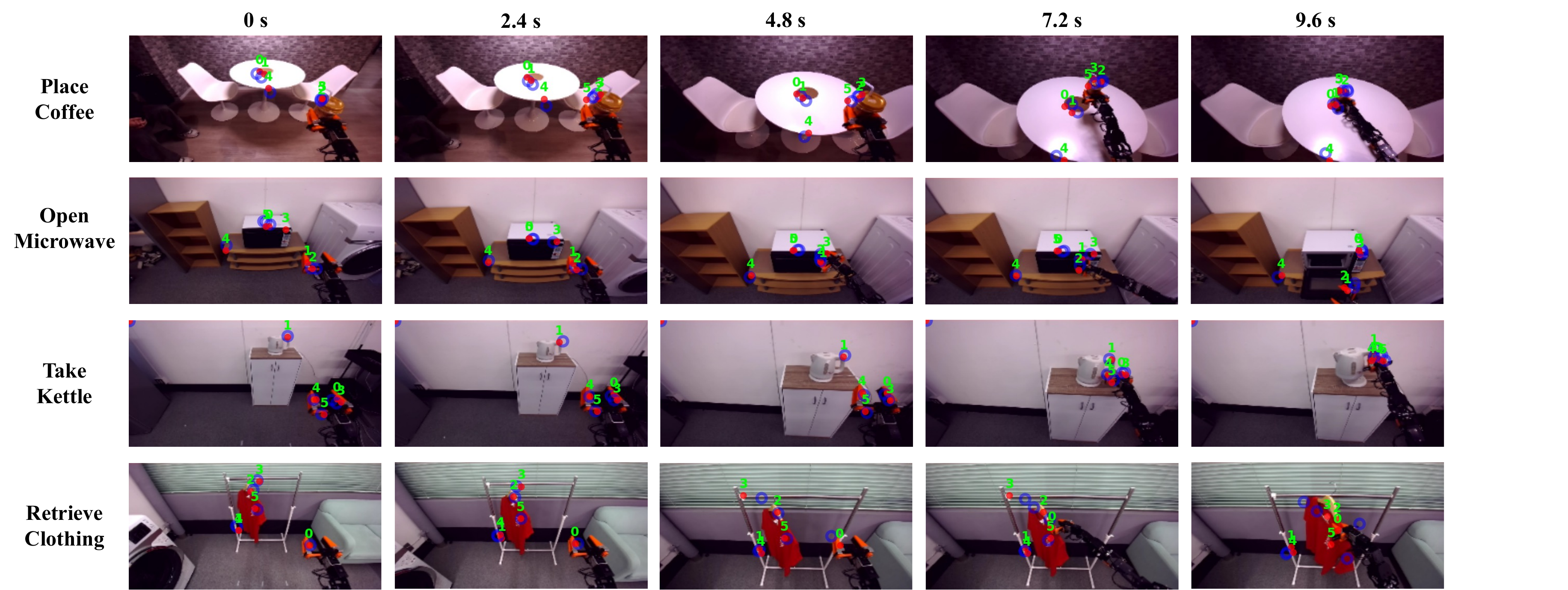}
    \caption{Visualization of attention points on the proposed model (left view). Red dots are extracted points $P_t$, blue circles are predicted points $\hat{P}_{t+1}$, and green numbers indicate output channels.}
    \label{fig:tasks_attentionpoints}
\end{figure*}

\begin{table*}
\centering
\caption{Motion Success Rates for Different Models}
\resizebox{\textwidth}{!}{
\begin{tabular}{l c c cc cc cc cc c}
\toprule
\multirow{2}{*}{\textbf{Model}} &
\multirow{2}{*}{\begin{tabular}[c]{@{}c@{}}\textbf{Input}\\\textbf{Image}\end{tabular}} &
\multirow{2}{*}{\begin{tabular}[c]{@{}c@{}}\textbf{Visual Encoder}\\\textbf{Backbone}\end{tabular}} &
\multicolumn{2}{c}{\textbf{Place Coffee}} &
\multicolumn{2}{c}{\textbf{Open Microwave}} &
\multicolumn{2}{c}{\textbf{Take Kettle}} &
\multicolumn{2}{c}{\textbf{Retrieve Clothing}} &
\multirow{2}{*}{\textbf{Avg. (99\% CI)}} \\
\cmidrule(lr){4-5} \cmidrule(lr){6-7} \cmidrule(lr){8-9} \cmidrule(lr){10-11}
 & & &
Coarse &
Precise (99\% CI) &
Insert &
Pull (99\% CI) &
Grasp &
Lift (99\% CI) &
Grasp &
Take (99\% CI) &
\\
\midrule
ACT & Stereo & ResNet
 & 62.0\%
 & 48.0\% (31.1--65.3)
 & 88.0\%
 & 74.0\% (55.9--86.5)
 & 22.0\%
 & 6.0\% (1.5--20.8)
 & 58.0\%
 & 56.0\% (38.3--72.3)
 & 46.0\% (37.2--55.1) \\

DP & Stereo & ResNet
 & 46.0\%
 & 42.0\% (26.0--59.9)
 & 22.0\%
 & 14.0\% (5.6--30.8)
 & 52.0\%
 & 4.0\% (0.8--18.0)
 & 60.0\%
 & 54.0\% (36.5--70.6)
 & 28.5\% (21.1--37.3) \\

$\pi0$ (3.5B) & Stereo & ViT
 & 40.0\%
 & 16.0\% (6.8--33.1)
 & 32.0\%
 & 24.0\% (12.1--42.0)
 & 0.0\%
 & 0.0\% (0.0--11.7)
 & 76.0\%
 & 76.0\% (58.0--87.9)
 & 29.0\% (21.5--37.8) \\

SmolVLA (0.45B) & Stereo & ViT
 & 22.0\%
 & 18.0\% (8.1--35.4)
 & 0.0\%
 & 0.0\% (0.0--11.7)
 & 0.0\%
 & 0.0\% (0.0--11.7)
 & 32.0\%
 & 32.0\% (18.0--50.2)
 & 12.5\% (7.7--19.8) \\

MSARNN & Stereo & SA
 & 22.0\%
 & 8.0\% (2.4--23.4)
 & 44.0\%
 & 44.0\% (27.7--61.7)
 & 76.0\%
 & 70.0\% (51.8--83.5)
 & 28.0\%
 & 28.0\% (15.0--46.2)
 & 37.5\% (29.2--46.6) \\

MSARNN & Mono & MSA
 & 40.0\%
 & 18.0\% (8.1--35.4)
 & 70.0\%
 & 70.0\% (51.8--83.5)
 & 0.0\%
 & 0.0\% (0.0--11.7)
 & 48.0\%
 & 44.0\% (27.7--61.7)
 & 33.0\% (25.1--42.0) \\

\textbf{MSARNN (Ours)} & \textbf{Stereo} & \textbf{MSA}
 & \textbf{90.0\%}
 & \textbf{72.0\% (53.8--85.0)}
 & \textbf{98.0\%}
 & \textbf{98.0\% (85.0--99.8)}
 & \textbf{96.0\%}
 & \textbf{92.0\% (76.6--97.6)}
 & \textbf{84.0\%}
 & \textbf{78.0\% (60.2--89.3)}
 & \textbf{85.0\% (77.4--90.4)} \\

\bottomrule
\end{tabular}
}
\label{table:motion_success_rate_combined_ci}
\end{table*}

We compare the proposed model with ACT, DP, SmolVLA (0.45B), and $\pi0$ (3.5B). To evaluate the stereo MSA contribution, we conduct two ablations: (1) replacing MSA with single-stage spatial attention (SA), and (2) a monocular MSARNN using only the left image. We deploy these methods using the LeRobot framework\cite{cadene2024lerobot}. All baseline methods are trained using their officially recommended configurations and hyperparameters. All methods are trained or fine-tune ($\pi0$: action expert and projections only) on the same dataset as setting showing Table \ref{tab:training_settings}. The validation performance curves confirm that all methods reach plateau. At each control step, a single policy query is performed, producing either one action or an action sequence (chunk/horizon). For sequence-based models (ACT/DP/SmolVLA/$\pi0$), only the first action is executed and re-planning occurs at the next step (receding-horizon), standardizing actuation frequency and preventing advantages from multi-step rollout or sampling. All methods use identical sensory inputs, preprocessing, and a single-step observation history. The proposed model, its ablations, and ACT are executed on the robot’s onboard CPU (Intel i7-1360p). Due to their computational requirements, DP, SmolVLA, and $\pi_0$ are executed on an RTX 4090-equipped desktop directly connected to the robot. Despite this difference in hardware, robot control frequency is strictly maintained at 10 Hz for all methods, and latency is reported per forward pass to reflect practical deployment conditions.

\section{Results and Discussion}

\subsection{Attention Points and Motion Generation}
For the proposed model, attention points are crucial for task execution, Fig.~\ref{fig:tasks_attentionpoints} shows the six pairs of attention points extracted and predicted by the model. The red dots represent the attention points extracted by the MSA module and the blue circles represent the attention points predicted by the motion predict module. These points consistently track task-relevant objects throughout the task, suggest that the learned attention representation aligns well with task progression. 

Table \ref{table:motion_success_rate_combined_ci} lists the final motion success rates with the full 99\% Wilson confidence interval bounds. Compared with ACT, DP, $\pi$0, and SmolVLA, the proposed method achieves the highest average success rate of 85.0\% under the evaluated conditions. The improvement is consistent across tasks with different characteristics, including rigid-object placement (Place Coffee), articulated-object manipulation (Open Microwave), grasping with orientation variation (Take Kettle), and deformable-object interaction (Retrieve Clothing). This improvement is particularly noticeable in tasks such as Open Microwave and Take Kettle under the evaluated settings.

\begin{figure}[!t]
    \centering
    \includegraphics[width=0.48\textwidth]{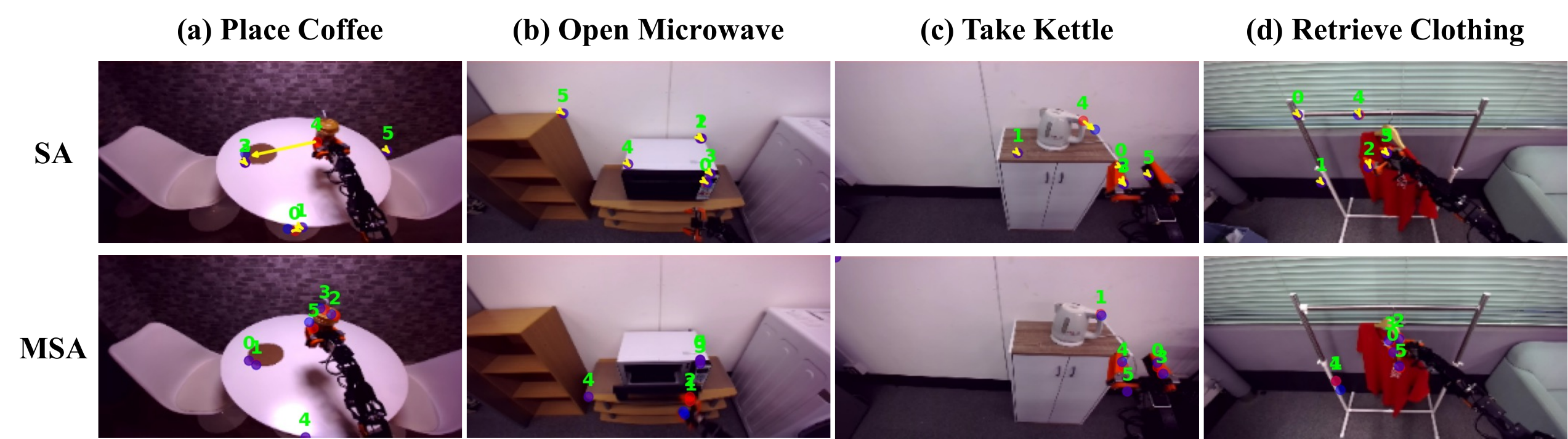}
    \caption{Attention points extracted by MSARNN with SA and MSA backbones. Red and blue dots denote points $P_t$ and ${P}_{t+1}$, respectively, green numbers indicate output channels, yellow arrows show the transition from $P_t$ to ${P}_{t+1}$.
}
    \label{fig:sa_attentionpoints}
\end{figure}

\begin{figure}[!t]
    \centering
    \includegraphics[width=0.43\textwidth]{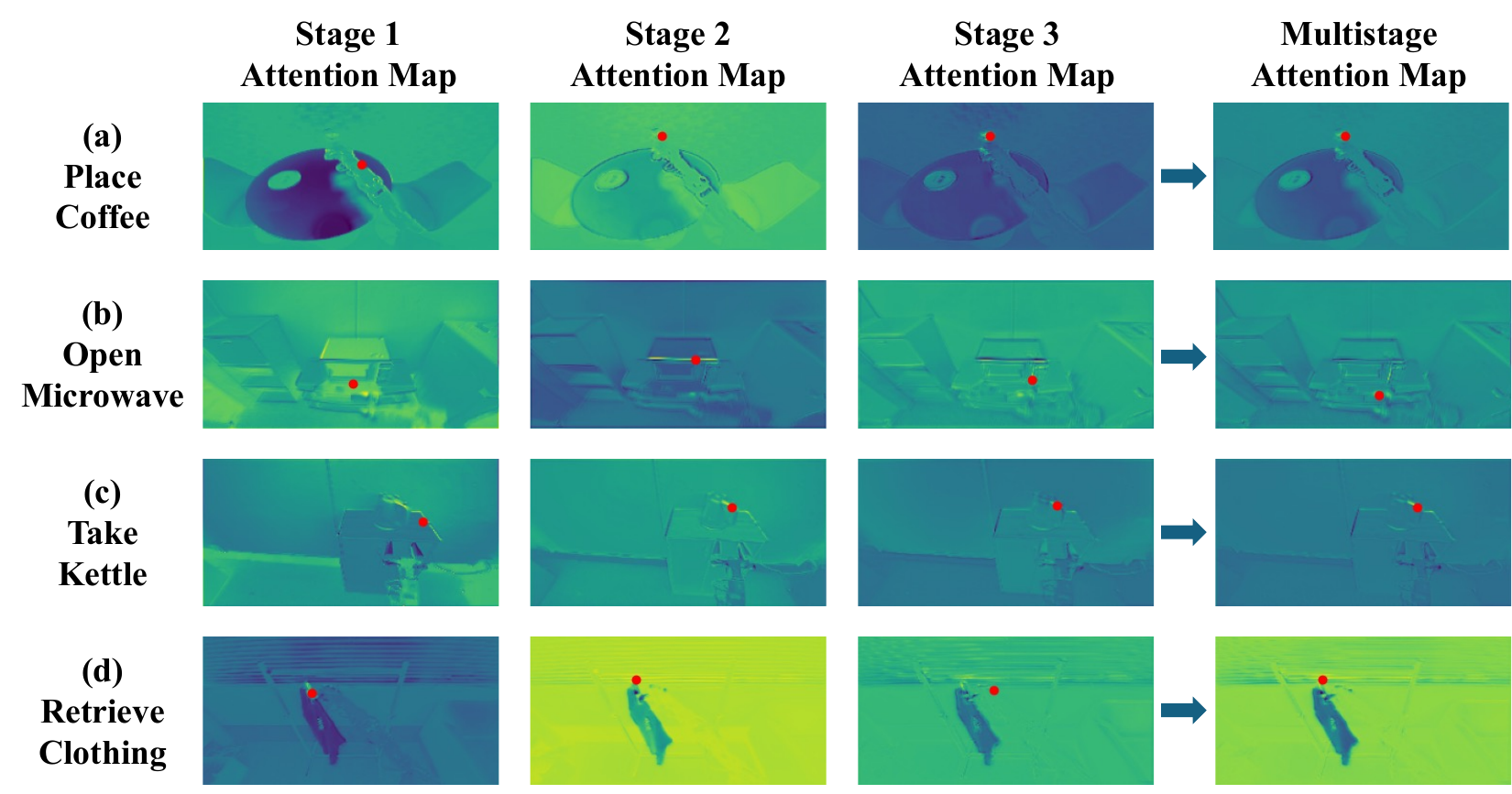}
    \caption{Visualization of attention maps within MSA during task execution. The red dots represent the highest value of the attention map.}
    \label{fig:msa_attentionpoints}
\end{figure}

\begin{figure*}[!t]
    \centering
    \includegraphics[width=0.98\textwidth]{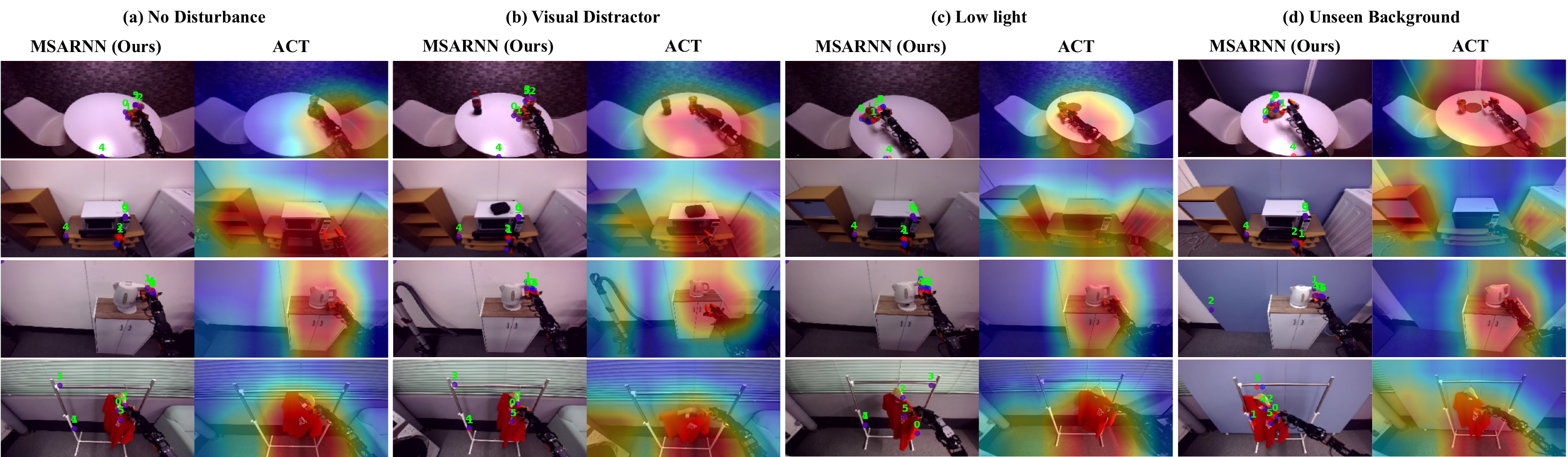}
    \caption{Attention representations comparison across models under different visual disturbance conditions. For MSARNN (MSA), red and blue dots denote attention points $P_t$ and ${P}_{t+1}$, with green numbers indicating output channels. For ACT (ResNet), attention intensity increases from blue to red.
}
    \label{fig:resnet_features}
\end{figure*}

\begin{table*}[!t]
\centering
\caption{Motion Success Rates Across Visual Disturbance Conditions}
\label{table:visual_disturbance}
\resizebox{0.75\textwidth}{!}{
\begin{tabular}{l l
                S[table-format=3.1] @{\%\ } l
                S[table-format=3.1] @{\%\ } l
                S[table-format=3.1] @{\%\ } l}
\toprule
\multirow{2}{*}{\textbf{Task}} &
\multirow{2}{*}{\textbf{Condition}} &
\multicolumn{2}{c}{\textbf{ACT}} &
\multicolumn{2}{c}{\textbf{ACT+AP (from MSARNN)}} &
\multicolumn{2}{c}{\textbf{MSARNN (Ours)}} \\
\cmidrule(lr){3-4} \cmidrule(lr){5-6} \cmidrule(lr){7-8}
 & &
\multicolumn{2}{c}{Success Rate (k/n, 99\% CI)} &
\multicolumn{2}{c}{Success Rate (k/n, 99\% CI)} &
\multicolumn{2}{c}{Success Rate (k/n, 99\% CI)} \\
\midrule
\multirow{4}{*}{Place Coffee}
 & No Disturbance
 & 48.0 & (24/50, 31.1--65.3)
 & 40.0 & (20/50, 24.4--58.0)
 & \bfseries 72.0 & \bfseries (36/50, 53.8--85.0) \\
 & Visual Distractor
 & 26.7 & (8/30, 11.6--50.2)
 & 36.7 & (11/30, 18.4--59.7)
 & \bfseries 66.7 & \bfseries (20/30, 43.4--83.9) \\
 & Low Light
 & 0.0 & (0/30, 0.0--18.1)
 & 20.0 & (6/30, 7.6--43.3)
 & \bfseries 36.7 & \bfseries (11/30, 18.4--59.7) \\
 & Unseen Background
 & 6.7 & (2/30, 1.3--27.7)
 & 13.3 & (4/30, 4.1--35.9)
 & \bfseries 40.0 & \bfseries (12/30, 20.9--62.7) \\
\midrule
\multirow{4}{*}{Open Microwave}
 & No Disturbance
 & 74.0 & (37/50, 55.9--86.5)
 & 76.0 & (38/50, 58.0--87.9)
 & \bfseries 98.0 & \bfseries (49/50, 85.0--99.8) \\
 & Visual Distractor
 & 6.7 & (2/30, 1.3--27.7)
 & 73.3 & (22/30, 49.8--88.4)
 & \bfseries 100.0 & \bfseries (30/30, 78.9--100.0) \\
 & Low Light
 & 26.7 & (8/30, 11.6--50.2)
 & 40.0 & (12/30, 20.9--62.7)
 & \bfseries 86.7 & \bfseries (26/30, 64.1--96.0) \\
 & Unseen Background
 & 3.3 & (1/30, 0.4--22.1)
 & 76.7 & (23/30, 53.2--90.5)
 & \bfseries 93.3 & \bfseries (28/30, 72.3--98.9) \\
\midrule
\multirow{4}{*}{Take Kettle}
 & No Disturbance
 & 6.0 & (3/50, 1.5--20.8)
 & 4.0 & (2/50, 0.8--18.0)
 & \bfseries 92.0 & \bfseries (46/50, 76.6--97.7) \\
 & Visual Distractor
 & 0.0 & (0/30, 0.0--21.1)
 & 3.3 & (1/30, 0.4--23.2)
 & \bfseries 96.7 & \bfseries (29/30, 78.7--99.5) \\
 & Low Light
 & 6.7 & (2/30, 1.3--27.7)
 & 10.0 & (3/30, 2.6--31.9)
 & \bfseries 96.7 & \bfseries (29/30, 78.7--99.5) \\
 & Unseen Background
 & 6.7 & (2/30, 1.3--27.7)
 & 3.3 & (1/30, 0.4--23.2)
 & \bfseries 43.3 & \bfseries (13/30, 23.9--66.1) \\
\midrule
\multirow{4}{*}{Retrieve Clothing}
 & No Disturbance
 & 56.0 & (28/50, 38.3--72.3)
 & \bfseries 78.0 & \bfseries (39/50, 60.2--89.3)
 & \bfseries 78.0 & \bfseries (39/50, 60.2--89.3) \\
 & Visual Distractor
 & 16.7 & (5/30, 7.3--35.8)
 & 33.3 & (10/30, 16.1--56.6)
 & \bfseries 80.0 & \bfseries (24/30, 57.3--92.7) \\
 & Low Light
 & 50.0 & (15/30, 30.8--69.2)
 & \bfseries 73.3 & \bfseries (22/30, 49.8--88.4)
 & 70.0 & (21/30, 47.8--85.5) \\
 & Unseen Background
 & 6.7 & (2/30, 1.3--27.7)
 & 26.7 & (8/30, 11.6--50.2)
 & \bfseries 56.7 & \bfseries (17/30, 35.5--75.4) \\
\midrule
\textbf{Overall Avg.}
 &
 & 24.8 & (139/560, 20.4--29.8)
 & 39.6 & (222/560, 34.5--45.1)
 & \bfseries 76.8 & \bfseries (430/560, 71.9--81.1) \\
\bottomrule
\end{tabular}
}
\end{table*}

To evaluate the necessity of stereo vision, we compare the proposed stereo MSARNN with a monocular ablation model, where the right camera image is replaced by a copy of the left image while keeping the network architecture, parameters, and training conditions unchanged \cite{xia2024dense}. As shown in Table \ref{table:motion_success_rate_combined_ci}, the monocular MSARNN shows performance degradation compared with the stereo version, particularly in tasks requiring accurate spatial perception. The average success rate drops from 85.0\% (Stereo) to 33.0\% (Mono), which suggest that stereo observations contribute to improved spatial consistency under the evaluated tasks. 

We further compare our method against previous spatial attention mechanisms. MSARNN (SA), employing a single-stage spatial attention, achieves an average success rate of 37.5\%, which is significantly lower than that of MSARNN (MSA). Although MSARNN (SA) performs reasonably in some tasks (e.g., Take Kettle), its performance is unstable across tasks. While it can identify task-relevant objects at times, the attention points drift as the task progresses, failing to maintain consistent focus. In some cases, it fails to fully capture the attention focus of task-relevant objects (Fig.~\ref{fig:sa_attentionpoints}). In contrast, the proposed model exhibits stable tracking performance. Fig.~\ref{fig:msa_attentionpoints} shows the attention maps from three selected attention heads of MSA module at three different stages of the attention map during task execution. The highest value in each attention map is marked with a red dot. Although not all attention maps consistently focus on the same object, multistage feature fusion ensures that final attention map converges toward a stable and consistent object focus, which improves robustness. These results suggest that the proposed multistage attention appears to capture task-relevant features more consistently at multiple stages and integrates them into robust attention points, leading to improved spatial cognition.

\subsection{Performance under Visual Disturbance}
We evaluate the robustness of the proposed model and ACT to visual disturbance under three conditions (Fig.~\ref{fig:resnet_features}). (i) Visual Distractor: introducing unseen objects or moving object around the target object, (ii) Low Light: reducing the indoor illumination to half of its original level, and (iii) Unseen Background: placing a blue background board. Each scenario is tested 30 times. Table \ref{table:visual_disturbance} lists motion success rates with the full 99\% Wilson confidence interval bounds. MSARNN maintains comparatively higher motion success rates than ACT across the evaluated disturbance conditions, with an overall average of 76.8\%. In contrast, ACT suffers significant performance drops under these visual disturbance.

We visualize the attention representations obtained by MSARNN and ACT (Fig.~\ref{fig:resnet_features}). ACT's attention visualization is derived from feature maps produced pre-trained ResNet~\cite{he2016deep}. To visualize the overall attention distribution, the 512-channel feature maps are averaged across channels to obtain a attention map. For MSARNN, the attention points remain spatially concentrated on task-relevant object across all evaluated conditions. In contrast, ACT exhibits more diffuse and less stable attention distributions for comparison with disturbance, especially under visual distractor and unseen background conditions. The attention heatmaps tend to spread over large areas, including background regions irrelevant to the manipulation target. These results suggest that ACT is more sensitive to visual disturbances, which negatively impact the motion generation.

We further evaluate the effect of stable attention points on ACT by integrating a pre-trained MSA module. As shown in Table \ref{table:visual_disturbance}, attention points increase the average success rate from 24.8\% to 39.6\%, which suggest that their benefit for ACT. We found that End-to-end training fails to extract stable attention points. One possible explanation is that ACT’s one-step sampling strategy may limit the extraction of temporally consistent attention features, preventing stable attention point extraction as in RNN-based models with full BPTT. Furthermore, we maintain that the ResNet feature outputs are the major factor affecting performance.

\subsection{Performance under Varying Initial Distances}
\begin{figure}[!t]
    \centering
    \includegraphics[width=0.45\textwidth]{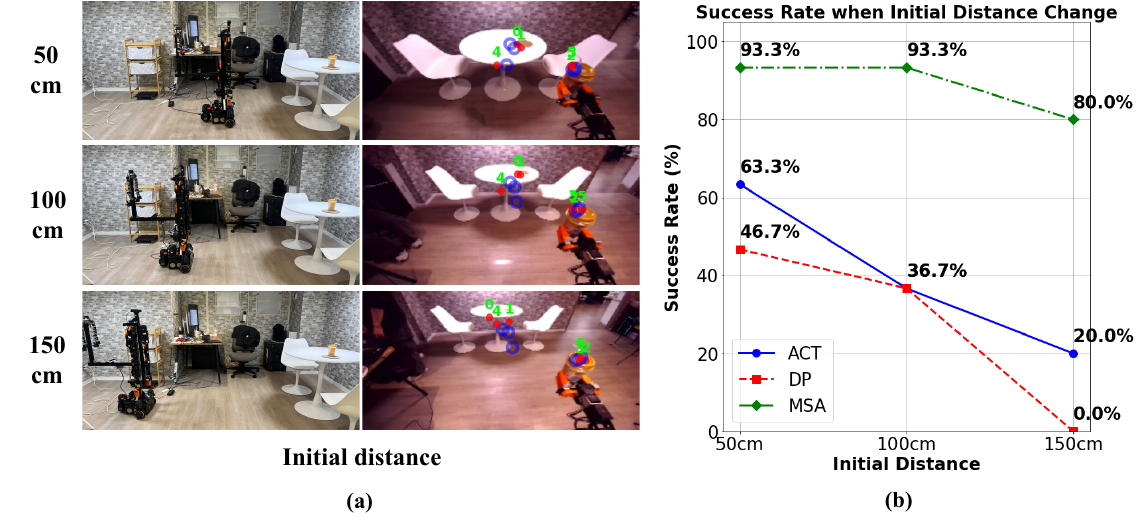}
    \caption{Results for different initial distances : (a) visualization of attention points on the proposed model, (b) motion success rates for different models.}
    \label{fig:initial_change}
\end{figure}

We evaluate performance under varying initial distances on the Place Coffee task by comparing the proposed method with ACT and DP. Models are trained at an initial distance of 50 cm and tested at unseen distances of 100 cm and 150 cm. Each condition is evaluated over 30 trials (20 trials from taught, 10 trials from unseen center positions). Fig.~\ref{fig:initial_change}(a) shows that, despite reduced object visibility at larger distances, the proposed model progressively localizes attention and achieves accurate placement. MSARNN maintain near-100\% success at 100 cm, with a minor degradation at 150 cm, significantly higher than ACT and DP (Fig.~\ref{fig:initial_change}(b)). These results suggest that the learned spatial representation maintains stable performance within the tested range of initial distances.

\begin{figure}[!t]
    \centering
    \includegraphics[width=0.45\textwidth]{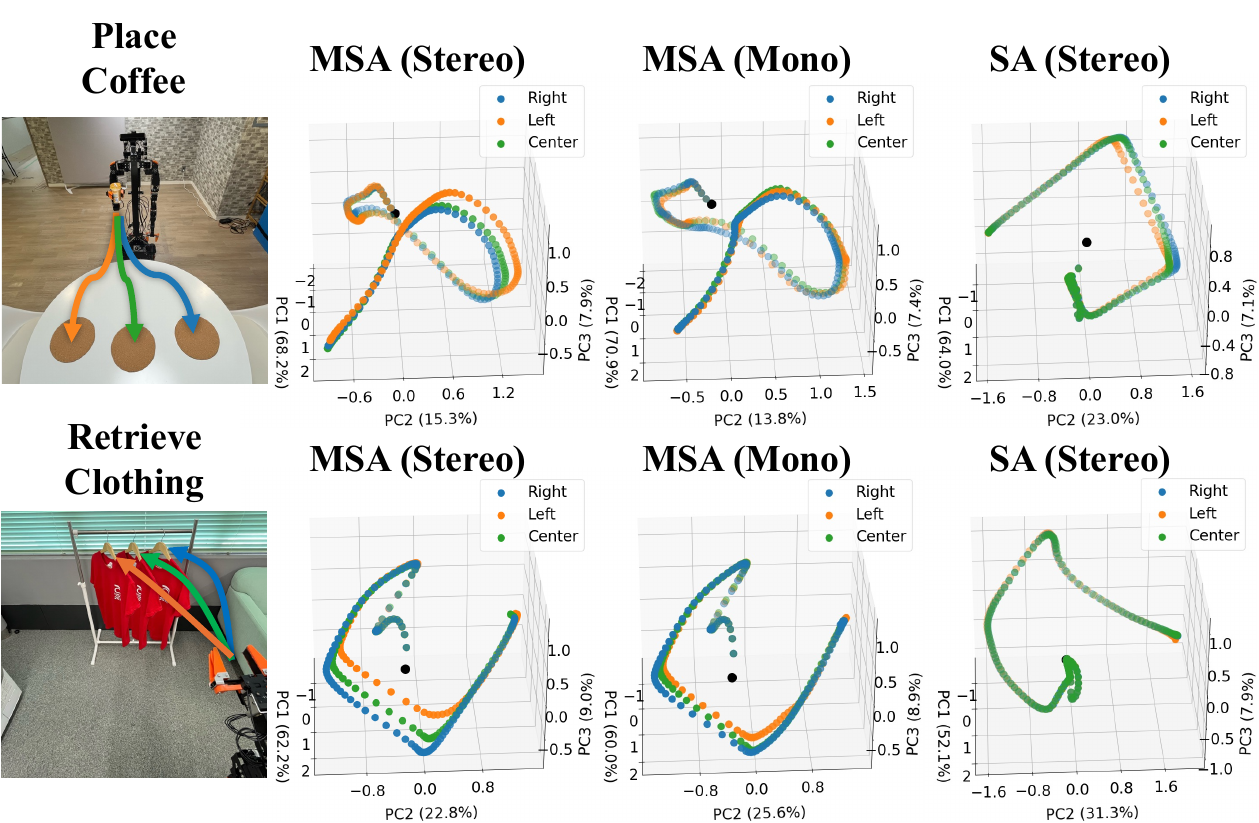}
    \caption{Visualizaion of  the temporal transitions of joint's low-level LSTM hidden states under MSARNN with stereo MSA, monocular MSA, and stereo SA backbones. Lines show state trajectories, with black dots marking initial states.}
    \label{fig:pca_msarnn}
\end{figure}

\subsection{Internal Representation Analysis}
To analyze how spatial information is encoded, we visualize the hidden states of the low-level joint LSTM using principal component analysis (PCA), as shown in Fig.~\ref{fig:pca_msarnn}. We compare the hidden states corresponding to different object positions in the Place Coffee and Retrieve Clothing tasks. The black dots indicate the hidden states at the start of each task. For the proposed stereo MSARNN, the hidden states form structured clusters that smoothly change with object position. For unseen center positions, the hidden states lie between those of the left and right positions. In contrast, the ablation models exhibit more entangled distributions with weaker correlation to object position. This suggests that the stereo MSARNN can utilize disparity information to encode task-relevant spatial information into its internal state, which contributes to improved tasks success rates.

\begin{table}
\centering
\caption{Robot Movement Distance (cm) Across Different Tasks \\ and Inference Latency}
\resizebox{0.48\textwidth}{!}{
\label{table:pose_accuracy_inference_latency}
\begin{tabular}{lcccccc}
\toprule
\textbf{Model} &
\shortstack{\textbf{Place}\\\textbf{Coffee}} &
\shortstack{\textbf{Open}\\\textbf{Microwave}} &
\shortstack{\textbf{Take}\\\textbf{Kettle}} &
\shortstack{\textbf{Retrieve }\\\textbf{Clothing}} &
\shortstack{\textbf{Inference}\\\textbf{Device}} &
\shortstack{\textbf{Latency}\\\textbf{(ms)}} \\
\midrule
ACT & 46.8$\pm$3.3 & 37.3$\pm$1.2 & 49.8$\pm$1.3 & 52.1$\pm$1.2 & CPU & 90.7$\pm$18.9\\
DP & 37.5$\pm$1.5 & 35.8$\pm$3.9 & 43.2$\pm$1.3 & 43.9$\pm$1.9 & GPU & 188.1$\pm$14.7 \\
$\pi0$ (3.5B) & 35.2$\pm$12.6 & 49.3$\pm$0.3 & 49.7$\pm$0.1 & 49.9$\pm$0.7 & GPU & 85.6$\pm$0.7 \\
SmolVLA (0.45B) & 38.9$\pm$4.1 & 26.5$\pm$2.1 & 42.3$\pm$2.3 & 39.3$\pm$6.6 & GPU & 90.4$\pm$2.1 \\
MSARNN (SA) & 57.7$\pm$6.8 & 43.2$\pm$1.0 & 49.3$\pm$0.7 & 52.2$\pm$0.3 & CPU & 22.8$\pm$13.5\\
MSARNN (Mono) & 55.0$\pm$6.9 & 49.6$\pm$2.2 & 53.6$\pm$0.9 & 49.3$\pm$0.8 & CPU & 33.3$\pm$12.1 \\
\textbf{MSARNN (Ours)} & 44.4$\pm$1.5 & 49.5$\pm$1.1 & 51.4$\pm$0.3 & 50.3$\pm$1.0 & CPU & \textbf{33.3$\pm$12.1}\\
\midrule
\textit{Dataset} & 50.6$\pm$2.4 & 48.1$\pm$2.7 & 48.7$\pm$1.8 & 50.5$\pm$1.3 \\
\bottomrule
\end{tabular}
}
\end{table}

\subsection{Motion Efficiency and Real-Time Performance}
We analyze motion efficiency by measuring the total robot movement distance, as reported in Table~\ref{table:pose_accuracy_inference_latency}. MSARNN (Ours) achieves movement distances close to the dataset average in most tasks. Certain models fail to accurately estimate the target object's position, leading to premature arm movements. For example, $\pi$0 in Place Coffee, SmolVLA in Open Microwave, as well as the DP and SmolVLA in Take Kettle exhibit noticeably shorter movement distances, which correlate with lower task success rates. Table \ref{table:pose_accuracy_inference_latency} further shows the end-to end inference latency. MSARNN (MSA) runs at a mean latency of 33.3 ms on the robot CPU, which is faster than other baseline models, while achieving higher success rates. These results suggest the feasibility of the proposed method for real-time control.

\section{CONCLUSIONS}
We propose an end-to-end motion generation method with stereo multistage spatial attention and hierarchical LSTM for mobile manipulation. While the current study improved robustness on controlled mobile manipulation tasks under moderate variations, further evaluation across broader scene distributions is left for future work. Future work will explore extending the framework to broader scene variations and investigating how spatial attention can be integrated with transformer-based policies for scalable learning.

\addtolength{\textheight}{-0pt}   




\section*{ACKNOWLEDGMENT}
This work was supported by the JST Moonshot Research and Development Project (JPMJMS2031) and the Research Institute of Science and Engineering, Waseda University. We would like to express our gratitude for this support.

\bibliographystyle{IEEEtran}  
\bibliography{refs} 

\end{document}